\documentclass[conference]{IEEEtran}

\IEEEoverridecommandlockouts
\usepackage{algorithm}
\usepackage{url}
\usepackage{multirow,makecell}
\usepackage{diagbox,booktabs,mdwlist}
\usepackage{cite}
\usepackage{amsmath,amssymb,amsfonts}
\usepackage{algorithmic}
\usepackage{graphicx}
\usepackage{textcomp}
\usepackage{xcolor}
\def\BibTeX{{\rm B\kern-.05em{\sc i\kern-.025em b}\kern-.08em
    T\kern-.1667em\lower.7ex\hbox{E}\kern-.125emX}}
\usepackage[colorlinks, urlcolor=cyan, linkcolor=red, anchorcolor=green, citecolor=blue]{hyperref}
\newcommand{\etal}{\textsl{et~al.}}

\begin{document}

\title{Radio Galaxy Morphology Generation Using DNN Autoencoder and Gaussian Mixture Models}

\author{\IEEEauthorblockN{Zhixian Ma$^{1}$,
  Jie Zhu$^{1}$, Weitian Li$^{2}$, Haiguang Xu$^{2}$}
  \IEEEauthorblockA{$^1$
   Department of Electronic Engineering, Shanghai Jiao Tong University, Shanghai, China
  }
  \IEEEauthorblockA{$^2$
  Department of Astrophysics, Shanghai Jiao Tong University, Shanghai, China}
  Email: \{mazhixian, zhujie\}@sjtu.edu.cn
}

\maketitle

\begin{abstract}
The morphology of a radio galaxy is highly affected by its central active galactic nuclei (AGN), which is studied to reveal the evolution of the super massive black hole (SMBH). In this work, we propose a morphology generation framework for two typical radio galaxies namely Fanaroff-Riley type-I (FRI) and type-II (FRII) with deep neural network based autoencoder (DNNAE) and Gaussian mixture models (GMMs). The encoder and decoder subnets in the DNNAE are symmetric aside a fully-connected layer namely code layer hosting the extracted feature vectors. By randomly generating the feature vectors later with a three-component Gaussian Mixture models,  new FRI or FRII radio galaxy morphologies are simulated. Experiments were demonstrated on real radio galaxy images, where we discussed the length of feature vectors, selection of lost functions, and made comparisons on batch normalization and dropout techniques for training the network. The results suggest a high efficiency and performance of our morphology generation framework. Code is available at: \url{https://github.com/myinxd/dnnae-gmm}.
\end{abstract}

\begin{IEEEkeywords}
radio galaxy morphology, generation, deep neural network (DNN), autoencoder, Gaussian mixture model (GMM)
\end{IEEEkeywords}

\section{Introduction}
\label{Sec.Intro}
The motivation of radio galaxy (RG) morphology generation is twofold. One is to obtain an automatic generator for more radio galaxy samples with known labels. The morphology of a radio galaxy is highly related to its central active galactic nuclei (AGN), which usually hosts a super massive black hole~\cite{Fabian2012,Padovani2017}. For different morphologies,  by which the radio galaxies can be classified, the evolution and mechanism of the AGNs are different. In the data release 7 (DR7) release of the FIRST (Faint Images of the Radio Sky at Twenty centimeters) survey at 1.4 GHz~\cite{Becker1995,Best2012}, there are more than 9.4$\times 10^{5}$ RGs, yet only are several-thousand clearly classified and labeled manually~\cite{Capetti2017a,Capetti2017b, Aniyan2017}.  
The other one is to benefit the foreground removal task on the observations from the Square Kilometer Array (SKA), which aims to uncover what happened after the Big Bang from the redshifted very weak 21 cm HI signal~\cite{Koopmans2015}. As a kind of very bright foreground signal, the RGs should be removed from the images so as to detect the target 21 cm signal for further study~\cite{Chapman2016}. Detection and removal of the RGs rely on the morphology study, and it is obvious that the finer understanding of the morphology, the more completely eliminating of the radio galaxies.

For the radio galaxies, there are two typical types namely Fanaroff-Riley type-I (FRI) and type-II (FII), which are with different morphologies~\cite{Fanaroff1974}.  A typical FRI is composed of a bright core and one or two plume-like lobes extending from the core to the edge of the lobes with decaying luminosity, while  a FRII is usually with separated hotspots brighter than the core at the ends of the lobes.  Wilman \etal has tried to simulate the FR radio galaxy morphology with circular core and two extended elliptical lobes~\cite{Wilman2008}, which may assist the theoretic study of radio galaxies, but is not applicable for the foreground removing task on real observations.

To obtain more vivid RGs, a generation model can be designed and trained by existing labeled RG samples. Recently, the generative adversarial network (GAN) was proposed by Goodfellow \etal ~\cite{Goodfellow2014GAN}, which consists with two subnets (i.e., the generator and the discriminator) and can generate new samples by the generator ~\cite{Tom2017,Kwon2018}. However, there exists non-convergence problem for training the GAN~\cite{Salimans2016}, and it cannot generate morphology of specific type only with the randomly generated Gaussian distributed inputs.  The autoencoder (AE) is another generation model, which is also composed of two subnets, namely encoder and decoder~\cite{Goodfellow2016}.  The encoder subtracts and extracts the features of the samples like encoding, and the decoder reconstructs the samples from the features like decoding and simulates new samples of specific type with randomly generated feature vectors that obey specific distributions (e.g., mixture Gaussian model). 

\begin{figure*}[t]
\centering
\includegraphics[width=0.8\textwidth]{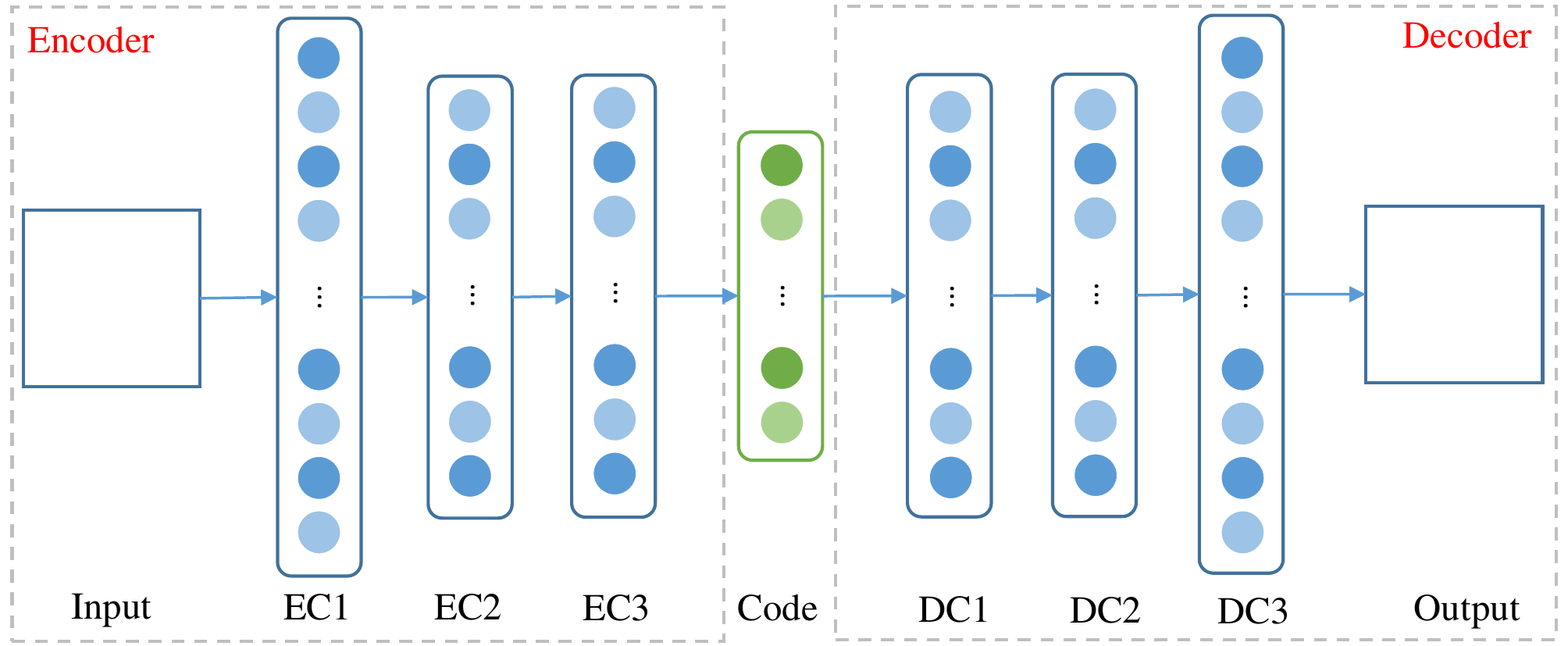}
\caption{Proposed DNN based autoencoder radio galaxy morphology generation network. EC and DC represents the layers of the encoder and decoder subnets, respectively.\label{fig.dnnae}} 
\end{figure*}

In this work, we propose a radio galaxy morphology generation framework. It takes advantage of the batch normalization (BN) layer for accelerating the network convergence~\cite{Ioffe2015} to form a deep neural network based autoencoder (DNNAE, see Fig.~\ref{fig.dnnae}).  Two three-component Gaussian mixture models are then estimated by the features extracted from the encoder, with which new feature vectors are randomly generated and input into the DNN decoder subnet for simulating new FRI/II radio galaxy morphologies.

This paper is organized as follows. In Sec.~\ref{sec.dnnae} we describe the proposed deep neural network based autoencoder and the training algorithm. In  Sec.~\ref{sec.gmm} the Gaussian mixture model for radio galaxy morphology generation is explained. Experiments are demonstrated and results are discussed in Sec.~\ref{sec.exp}. We conclude in Sec.~\ref{sec.con} with outlooks.

\section{Deep Neural Network Autoencoder}
\label{sec.dnnae}
We illustrate the proposed  deep neural network based autoencoder (DNNAE) in Fig.~\ref{fig.dnnae} and list the parameters setting in Table \ref{tab.dnnae}, which is composed of an encoder and a decoder. They are with symmetric structure aside a single layer namely code layer. Both of them are the DNN subnets composing of fully-connected (FC) layers. 

As introduced by Ioffe and Szegedy~\cite{Ioffe2015}, the batch normalization (BN) avoids distribution changing problem between connected layers, so as to accelerate the network training. In addition, there is no need for dropout since BN regularizes the network parameters. Therefore, batch normalization is processed on each FC layers in both of the encoder and decoder in this work. We will discuss and compare the performance of BN and dropout techniques on training the DNNAE networks with experiments in Sec.~\ref{sec.exp}.

To train the parameters in the DNNAE network,  a loss function (or cost function) should be defined.  Since our target is to generate the morphology of radio galaxies belonging to specific types (i.e., FRI or FRII), we focus on the features that contains singularities between the two types, and the similarities as well. 

As for the autoencoder, It usually applies the mean squared error (MSE) between the input of the encoder and the output of the decoder as the objective to be minimized~\cite{Bishop2006,Zhang2014}. By optimizing such object, the network tend to extract features both of the two RG types have. Another cost function namely cross-entropy (CE) is widely applied in classification tasks~\cite{Bishop2006,Bengio2007}, which is minimized for extracting the most distinguishable features between the samples of different types. The MSE and CE loss functions are defined as,
\begin{align}
\label{eq.mse}
L_{\rm{MSE}} &= \frac{1}{N_{\rm{B}}}\sum^{N_{\rm{B}}}_{i=1}\sum^{N_{\rm{row}}}_{j=1}\sum^{N_{\rm{col}}}_{k=1}{|I_{i,j,k} - O_{i,j,k}|^2}, \\
L_{\rm{CE}} &= \frac{1}{N_{\rm{B}}}\sum^{N_{\rm{B}}}_{i=1}\sum^{N_c}_{j=1}{-(y_{i,j}\cdot\log_2{p_{i,j}})},
\end{align}
where $L_{\rm{MSE}}$ is the mean squared error between the reconstructed images $O$ and original input images $I$. $N_{B}$ is the number of images in one batch, and $N_{\rm{row}}$ and $N_{\rm{col}}$ are the number of rows and columns of the images. $L_{\rm{CE}}$ is the cross entropy loss. $N_c$ represents number of types and is set as two. $y_{i, j}$ is the one-hot real label of the $j$th RG sample ($y_{i,j} = 1$ if the source belongs to type $j$, and $y_{i,j} = 0$ if otherwise).  $p_{i, j}$ is the normalized probability of this RG being classified as type $j$ in a certain batch.

To train our morphology generation network, a simple but efficient way is to make a combination of them as the objective to be optimzed, i.e.,
\begin{equation}
\label{eq.obj}
\min{L} = \min\limits_{\rm E, D}L_{\rm MSE} + \min\limits_{\rm E}L_{\rm CE},
\end{equation}
where $L$ is the combined loss, and E and D represent the encoder and decoder. In this work, parameters in the DNNAE network are trained with both the MSE and CE losses alternatively, where the CE loss is back-propagated with gradients to parameters in the encoder subnet, and the MSE loss is to the whole network. (see Alg.~\ref{alg.train} for details).

\begin{table}[t]
\renewcommand{\arraystretch}{1.2}
\caption{Parameters setting for the proposed DNN based autoencoder network. AF represents the active function and BN means batch normalization. Y and N are the flags for whether a batch normalization is appended to this layer.\label{tab.dnnae}}
\centering
\small
\begin{tabular}{|l|c|c|c|c|}
\hline
Subnet & layer & structure &  AF & BN\\
\Xhline{1pt}
\multirow{4}{*}{Encoder} & Input & $40\times40\times1$ & --- & ---\\ \cline{2-5}
&EC1 & 2048 & ReLU & Y\\ \cline{2-5}
&EC2 & 1024 & ReLU & Y\\ \cline{2-5}
&EC3 & 1024 & ReLU & Y\\ \Xhline{1pt}
&Code & 256 & ReLU & N \\ \Xhline{1pt}
\multirow{4}{*}{Decoder} &DC1 & 1024 & ReLU & Y\\ \cline{2-5}
&DC2 & 1024 & ReLU & Y\\ \cline{2-5}
&DC3 & 2048 & ReLU & Y\\ \cline{2-5}
&Output & $40\times40\times1$ & Sigmoid & ---\\
\hline
\end{tabular}
\end{table}

Some popular techniques are also applied for training the proposed DNNAE network. The rectified linear unit (ReLU~\cite{Glorot2011}) is applied as the activation function after each fully-connected layers, and we apply the adaptive moment optimization function (ADAM;~\cite{Kingma2014}) to adjust the parameters with exponentially decaying learning rates.

\begin{algorithm}[t]
  \caption{DNNAE-MSE+CE training algorithm.}
  \label{alg.train}
  \small
  \begin{algorithmic}[1]
    \STATE \textbf{Input:} samples $S$ and labels $L$ 
    \STATE \textbf{Input:} $epochs$ and $batchsize$
    \STATE batches = length(labels) / batchsize 
    \FOR {$i = 1 : epochs$}
    	\FOR {$j = 1 : batches$}
    	\STATE {$S_b = S[(j)*batchsize+1 : (j+1)*batchsize]$}
    	\STATE{$L_b = L[(j)*batchsize+1 : (j+1)*batchsize]$}
    	\STATE {Feedforward $S_b$ and $L_b$ to obtain $L_\mathrm{CE}$ and $L_\mathrm{MSE}$}
    	\STATE {Backpropogate $L_\mathrm{CE}$ to parameters in the encoder subnet}
    	\STATE {Backpropogate $L_\mathrm{MSE}$ to parameters in the whole net}
    	\ENDFOR
    \ENDFOR
  \end{algorithmic}
\end{algorithm}

\section{Generation Algorithm}
\label{sec.gmm}
By feeding randomly generated feature codes, which obey the distribution of the features extracted from the real radio galaxy samples, into the decoder subnet of the DNNAE can it output simulated new radio galaxy images.  We deploy a three-component Gaussian mixture models (GMM) to fit the distributions of the extracted FRI or FRII radio galaxy features, in which one component is for the similarities between the FRI/II types, and the others are for the singularities of them.

Denote $\mathbf{f} = \{f_1, f_2, ..., f_M\}$ as the feature vector (i.e., the code) and $M$ is the length of the features.  Then the three-component GMM is,
\begin{equation}
\label{eq.gmm}
P(\mathbf{f} | \mathbf{\theta}) = \sum^{K}_{k=1}{\alpha_k \phi(\mathbf{f}|\mathbf{\theta_k})},
\end{equation}
where $P(\mathbf{f} | \mathbf{\theta})$ represents the probability that the feature vector $\mathbf{f}$ is generated from this GMM. $\alpha_k \geq 0, \sum^K_{k=1}{\alpha_k} \equiv 1, k=1,2,...,K$ are the coefficients and $\mathbf{\theta} = \{\mathbf{\mu_k}, \Sigma_k\}$ are the parameters of the corresponding Gaussian models. $K$ is the number of components, which is set as three in this work, where two components are for the singularities of the FRI and FRII morphologies and the rest one for the similarities of them. $\phi(\mathbf{f}|\mathbf{\theta_k})$ is the $k$th Gaussian component that is defined as,
\begin{equation}
\phi(\mathbf{f}|\mathbf{\theta_k}) = \frac {1} {(2 \pi)^{M/2}| {\Sigma_k}|}\exp{[{- \frac{1}{2} (\mathbf{f} - \mathbf{\mu_k})^T {\Sigma_k}^{-1}(\mathbf{f} - \mathbf{\mu_k})}]}.
\end{equation}

For each type of the RGs, a GMM is constructed and estimated to obtain corresponding $\mathbf{\theta_k}$, which is later used to randomly generate new specific RG images. The expectation maximization algorithm is used to estimate the GMM parameters~\cite{Ma2017,Bilmes1998}. 

\section{Experiments and results}
\label{sec.exp}
To evaluate the performance of our proposed DNNAE network as well as the GMM based generation algorithm, we demonstrate experiments, and make discussions on the results in this section.  

\begin{figure*}[t]
\centering
\includegraphics[width=1.0\textwidth]{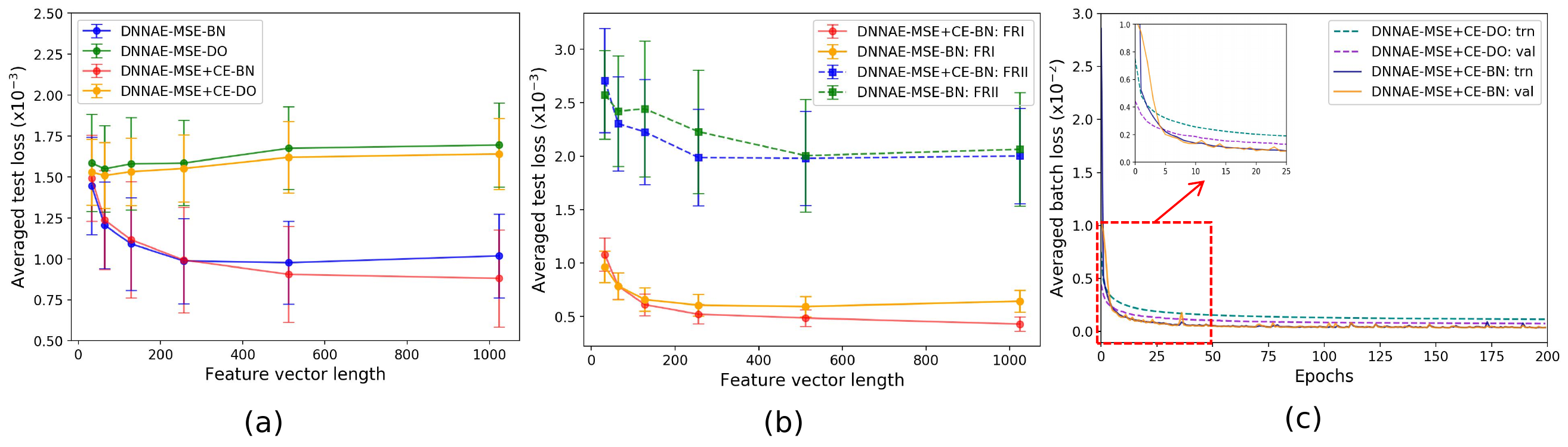}
\caption{Experimental results and comparisons. (a) are the averaged test losses of networks with different code feature lengths, loss functions, and dropout (DO) or batch-normalization (BN) techniques. (b) are the within-class test losses on FRI/II RGs with networks of different loss functions at variant code lengths. (c) are the training and validation loss curves of networks with fixed MSE+CE loss functions and 256 code length while with dropout or batch-normalization techniques.\label{fig.cmp} } 
\end{figure*}

\begin{figure*}[t]
\centering
\includegraphics[width=0.95\textwidth]{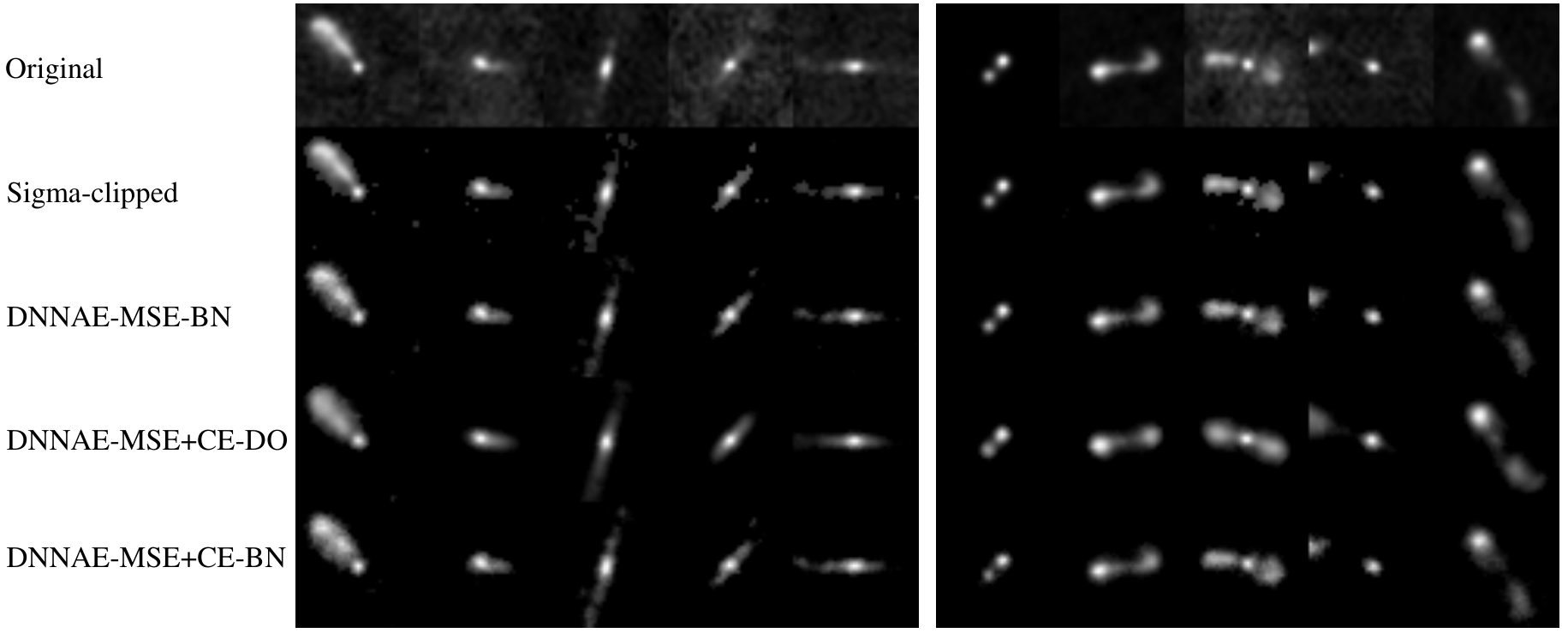}
\caption{Reconstructed radio galaxy images by DNNAE-MSE-BN, DNNAE-MSE+CE-DO, and DNNAE-MSE+CE-BN networks, the code length are 256 in all of them. Left and right five columns are FRIs and FRIIs, respectively\label{fig.cmp_recon}} 
\end{figure*}

\subsection{Data preparation}
Real radio galaxy images were selected from two catalogs, i.e., the FRICAT~\cite{Capetti2017a} and the FRIICAT~\cite{Capetti2017b}, to form the samples (192 FRIs and 99 FRIIs) for training the network, and were retrieved from the FIRST data archive\footnote{FIRST image cutouts: \url{https://third.ucllnl.org/cgi-bin/firstcutout}}. 

Before fed into the DNNAE network, the original images were preprocessed by three steps. First, the noise was suppressed using the sigma clipping algorithm~\cite{Aniyan2017} to improve the contrast of the radio galaxy morphologies. Second, the center region of 40$\times$40 pixels was cropped from the 150$\times$150 pixels image each. The last step was applying data augmentation to enlarge the sample numbers for avoiding overfitting and achieving a balanced training set. In this work, the cropped sample images were augmented by flipping (left-to-right, up-to-bottom, or diagonal) and rotated with uniformly distributed angle $\beta$ ($\beta\in[0, 360^\circ)$) . 

The 291 radio galaxy samples were randomly divided into training, validation and test subsets with a ratio of 64\% : 16\% : 20\% before augmentation. The FRI samples were 200 times augmented each (i.e., 24,600 for training and 6,200 for validation), and the FRIIs were 400 times augmented each (i.e., 23,200 for training and 6,400 for validation).  Note that the test samples (38 FRIs and 20 FRIIs) were not augmented. 

\subsection{Experiments and comparisons}
The proposed DNNAE was constructed as Fig.~\ref{fig.dnnae} illustrated, where length of the feature vectors varied for discussions. Since the selection of loss function affects the performance of the network, and to compare our MSE+CE loss function strategy, a group of networks with only MSE loss was also formed. We did not train a DNNAE with cross-entropy loss function, since this loss is only back-propagated to the encoder subnet instead of the whole net.  In addition, two networks that apply the dropout technique without batch normalization were also constructed for comparison. To be more intuitive, we name the four networks as DNNAE-MSE+CE-BN, DNNAE-MSE-BN, DNNAE-MSE+CE-DO, and DNNAE-MSE-DO, where BN and DO are the abbreviations of batch normalization and dropout. 

All the networks were batchly trained with the training and validation subsets above during 200 epochs, where the batch's size was set as 100. The exponentially decaying learning rate for parameters optimizing was initialized as 0.001 and varied by a decaying rate of 0.95. For the dropout applied networks, the keep probability was 0.5.

Experimental results and comparisons of the four networks are illustrated in Fig.~\ref{fig.cmp}. In Fig~\ref{fig.cmp}(a) shows the MSE losses of the test subset by the DNNAE-MSE+CE-BN, DNNAE-MSE-BN, DNNAE-MSE+CE-DO, and DNNAE-MSE-DO networks with variant feature vector length at the code layer, where the error bars are under a 95\% confidence level. In Fig.~\ref{fig.cmp}(b) we illustrate the within-class test loss of FRI/II RGs with the DNNAE-MSE+CE-BN and DNNAE-MSE-BN networks. And in Fig~\ref{fig.cmp}(c) the training and validation loss during 200 epochs between DNNAE-MSE+CE applying BN and dropout are compared, where the code length is fixed as 256. 

In addition, ten FRI and FRII test samples were randomly selected to evaluate the reconstruction performance of the DNNAE-MSE-BN, DNNAE-MSE+CE-DO, and DNNAE-MSE( see Fig.~\ref{fig.cmp_recon}).

From the experimental results, we summarize and discuss that, 
\begin{itemize}
\item
In general, the proposed DNNAE network can reconstruct FRI/II radio galaxies with low reconstruction error and high efficiency.
\item
For all the networks, the testing loss tend to converge as the code feature length increases. Especially from Fig.~\ref{fig.cmp}(b), the losses of FRI/II converge at code length of 256. That's why we select 256 as the code length. 
\item
From Fig.~\ref{fig.cmp}(a) and (b), it can be found that the combination of MSE and CE loss functions achieves better performance than the case only applying MSE loss.  
\item
From Fig.~\ref{fig.cmp}(c), it is obvious that batch normalization accelerates the convergence of network parameters. In addition, BN can avoid the saturation problem while the parameter space enlarges, see the curves of DNNAE-MSE+CE-DO and DNNAE-CE-DO in Fig.~\ref{fig.cmp}(a).  
\item
The test losses of FRIs are lower than the FRIIs. In our view, it is because the FRII morphologies are more complicated, and the shortage of FRII samples may also be a problem.
\item 
For the reconstructed images by the three networks, the DNNAE-MSE+CE-BN and DNNAE-MSE-BN achieve similar performance in visual, but the net of MSE+CE is better at some RGs on finer substructures, e.g. the FRIs at column one and four, and the FRII at column nine.  For the DNNAE-MSE+CE with dropout, it could not reconstruct some RGs, e.g., the FRI at column five and the FRII at column ten.
\end{itemize}

\subsection{Sample generation}
\begin{figure}[t]
\centering
\includegraphics[width=0.45\textwidth]{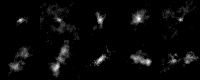}
\caption{Generated FRI (top row) and FRII (bottom row) radio galaxy morphologies by the DNNAE-MSE+CE-BN-256 and the three-component GMMs.\label{fig.sim}} 
\end{figure}

On the DNNAE-MSE+CE-BN net, we simulated images of FRI/II radio galaxy morphology using the three-component GMMs described in Sec.~\ref{sec.gmm}. Two GMMs were estimated from the features extracted by the network of the training and validation samples for the two RG types, respectively. The generated images were displayed in Fig.~\ref{fig.sim}, which achieved distinguishable morphologies between the FRI and FRII radio galaxies and were correctly classified by all of the authors in visual. 

\section{Conclusion}
\label{sec.con}
A deep neural network based autoencoder is proposed to generate radio galaxy morphologies, which combines the mean squared error (MSE) and cross entropy (CE) loss functions and applies the batch normalization training technique. To simulate specific FRIs and FRIIs, three-component Gaussian mixture models are estimated to randomly generate feature vectors that are fed into the decoder subnet to output new radio galaxy morphology samples.

Results of the experiments suggest that reconstruction loss of the network converges when the feature vector length increases. Compared with the network with only MSE loss, our MSE+CE combination strategy achieved better performance. The batch normalization technique made the network's parameters converge faster and achieved significant low reconstruction error. The extracted features by the DNNAE network could be well described by the Gaussian mixture models, for both the similarities and singularities of the RGs with different morphologies.

In the future, we will add more types of radio galaxies with complicated morphologies to train a more general generator model.

\section*{Acknowledgment}
This work is supported by the National Natural Science Foundation of China (grant Nos. 61371147 and 11433002), and the National Key Research and Discovery Plan (grant No.2017YFF0210903).


\end{document}